# LPBSA: Enhancing Optimization Efficiency through Learner Performance-based Behavior and Simulated Annealing

Dana Rasul Hamad[1] and Tarik A. Rashid[2*]
[1] Computer Science Department, Faculty of Science, Soran University, Soran, Erbil, KRG, Iraq. Email. dana.hamad@soran.edu.iq
[2] Computer Science and Engineering Department, University of Kurdistan Hewler, Erbil, KRG, Iraq. Corresponding email: tarik.ahmed@ukh.edu.krd

**Abstract**

This study introduces the LPBSA, an advanced optimization algorithm that combines Learner Performance-based Behavior (LPB) and Simulated Annealing (SA) in a hybrid approach. Emphasizing metaheuristics, the LPBSA addresses and mitigates the challenges associated with traditional LPB methodologies, enhancing convergence, robustness, and adaptability in solving complex optimization problems. Through extensive evaluations using benchmark test functions, the LPBSA demonstrates superior performance compared to LPB and competes favorably with established algorithms such as PSO, FDO, LEO, and GA. Real-world applications underscore the algorithm's promise, with LPBSA outperforming the LEO algorithm in two tested scenarios. Based on the study results many test function results such as TF5 by recording (4.76762333) and some other test functions provided in the result section prove that LPBSA outperforms popular algorithms. This research highlights the efficacy of a hybrid approach in the ongoing evolution of optimization algorithms, showcasing the LPBSA's capacity to navigate diverse optimization landscapes and contribute significantly to addressing intricate optimization challenges.

## 1. Introduction

There are many metaheuristic algorithms have been proposed to tackle intricate problems that have the advantage of swiftly providing approximate solutions, even for problems of considerable complexity [1]. Improving these algorithms became a significant remark by the researchers to achieve these algorithm purposes, and usually persuade the algorithm developers to observe the manifestation of their concern towards their inventions [2],[3]. Learner performance-based behavior (LPB) is one of those algorithms that has recently been developed, the idea behind this algorithm is that the concept involves admitting high school graduates to the university [4]. Modifying and improving this algorithm was one of the LPB developer's research directions. The research aims to provide insights into the strengths and weaknesses of the LPB algorithm, both in its original form and the enhanced version with Simulated Annealing. The goal is to contribute to the advancement of optimization algorithms tailored for guiding learners through educational pathways, with a focus on achieving a balance between exploration and exploitation for efficient and effective learning path planning. The LPB algorithm is designed to facilitate the learner's transition from high school to higher instructions like the university by refining their learning behavior. Also, the primary objective is to optimize the LPB algorithm's performance in terms of convergence, exploration, and exploitation, ultimately enhancing its efficiency in guiding learners.

Learner performance-based behavior using Simulated Annealing or LPBSA is an improved approach that will be more focused in the presented work. Optimization algorithms play a pivotal role in improving the efficiency of intelligent systems by fine-tuning parameters, reducing computational complexity, and enhancing overall performance. Notably, research by [5] highlights the importance of improving techniques in machine learning and neural network training and accuracy, demonstrating how advancements in optimization algorithms can significantly accelerate convergence and reduce training time. Many real-world problems involve dynamic environments where the optimal solution may change over time, Metaheuristic algorithms with adaptive mechanisms can dynamically adjust their search strategies, enabling them to the scope with changing conditions [6]. The ongoing improvement of these adaptive features ensures the relevance and applicability of modified algorithms. Improving an algorithm by using another algorithm could become a significant aspect of the researcher's direction, this method kinds are used in different styles such as hybridization, combination, and grouping algorithms, and they can advantage of real-world examples. Merging diverse optimization algorithms enhances the hybrid algorithm's effectiveness in both precision of calculations and computational efficiency when estimating the source term for hazardous gases, this outcome carries significant implications for emergency management in handling hazardous stations [7]. Simulated Annealing is a probabilistic technique or stochastic approach that has been used to improve LPB in the current work. Many related examples are available which will be focused on in the next section. The contribution of this research work can be outlined as follows:

- Improving LPB using Simulated Annealing
  1. Global Exploration: The integration of simulated annealing enhances LPB's exploration capabilities of suboptimal solutions, aiding in escaping local optima.
  2. Temperature-Controlled Evolution: Simulated annealing introduces temperature-controlled evolution, regulating the acceptance of suboptimal solutions. This controlled evolution helps overcome local optima, improving the algorithm's ability to reach more favorable regions.
- Combined contribution: The combination of LPB and SA results that achieves a synergistic effect:
  1. Adaptive Learning + Global Exploration: The combined approach of adaptive learning with simulated annealing's global exploration, providing a balanced and powerful optimization strategy.
  2. Synergistic Performance: The improved algorithm exhibits advantages in convergence solution, quality, and robustness across diverse problem domains, showcasing the synergy between LPB and SA.

The objectives of this study are outlined to address specific aspects related to the improvement of the Learning Path-Based (LPB) algorithm using Simulated Annealing (SA). The primary goals include:

1. Algorithm Enhancement: Performance by incorporating Simulated Annealing, Convergence, and Exploitation: the speed of convergence and exploitation capability enhancement of LPB.
2. Benchmark Evaluation: Test Function Evaluation: Employ a diverse set of benchmark functions, including unimodal, multi-modal, and composite functions, to thoroughly assess the algorithm's performance. Comparison with Other Algorithms: Compare the enhanced LPB algorithm against other optimization algorithms, Particle Swarm Optimization (PSO), Genetic Algorithm (GA), and Differential Evolution Algorithm (DA) an example of those optimization algorithms.

3. Exploration and Avoidance of Local Optima: Multi-Optima Handling: Evaluate the LPB-SA algorithm's ability to handle multi-modal functions with multiple optima. Local Optima Avoidance: Assess the algorithm's capacity to explore the search space effectively and avoid being trapped in local optima.
4. Efficient Learning Path Planning: Studying Behavior Improvement: Focus on improving the LPB algorithm's effectiveness in enhancing learners' studying behavior during the transition from high school to university. Balancing Exploration and Exploitation achievement to optimize the learning path planning process.
5. Comprehensive Experimentation: Parameter Tuning: Investigate the impact of population size, crossover, and mutation rate and divide population as the key parameters on the LPB-SA algorithm's performance. Simulated Annealing Contribution: Analyze the specific contributions of Simulated Annealing to the overall improvement of the LPB algorithm.

These objectives collectively aim to contribute valuable insights into the effectiveness of the LPB algorithm with Simulated Annealing and its potential for optimizing learning path planning in educational contexts. The rest of the papers are organized as follows: The introduction section sets the stage by presenting the context, objectives, and contributions of the study, focusing on optimizing learner path planning. Following this, the related work section reviews existing algorithms in learner path planning, highlighting their strengths and limitations. Section three details the Learner Performance Behavior (LPB) algorithm, elucidating its design principles and educational application. Simulated Annealing (SA) is explored in section four, offering a historical overview and perspectives from researchers. The fifth section introduces the Improved LPB using Simulated Annealing (LPB-SA), detailing the integration of SA and modifications to LPB. Section six rationalizes the choice of Simulated Annealing, while section seven presents empirical results and facilitates discussion. The paper concludes in section eight, summarizing contributions and discussing implications and future avenues for research in educational optimization algorithms.

## 2. Related Work

The LPB algorithm, drawing on genetic algorithms and influenced by Darwinian evolution, has proven effective in addressing optimization challenges through adaptive learning. Renowned for its ability to adapt dynamically to population performance, LPB offers a sturdy framework for optimization tasks. To strengthen the balance or the equilibrium between exploration and exploitation in LPB a hybrid method is incorporated by integrating simulated annealing (SA). Simulated annealing, drawing inspiration from metallurgical annealing procedures, introduces a global search strategy to supplement the local search capabilities of LPB. This fusion of LPB and SA is designed to collaboratively progress the overall of the algorithm's performance. The combination of algorithms has led to notable achievements in optimization. In [8], the Genetic Algorithm with Active Set Technique (GA-AST) enhanced the precision of estimating temperature profiles in the human head. [9][10] illustrated the effectiveness of merging the Genetic Algorithm with an internal Point Technique (GA-IPT) to solve the Painlevell equation. Similarly, [10] demonstrated how to use GA and IPT together to optimize a "feed-forward" neural network for solving the porous fin equation.

The neuro-heuristic approach involving GA and Sequential Quadratic Programming (GA-SQP), as presented in [11], affirmed the viability and efficacy of integrating genetic algorithms with other optimization techniques. In addition to LPB incorporating SA, numerous research studies have effectively merged simulated annealing with different algorithms, demonstrating the flexibility of SA in a hybrid context. In the study described in [12], a combination of simulated annealing and genetic algorithm (SA-GA) was utilized to enhance the optimization of mechanical structure designs. The hybrid approach exhibited better convergence rates and solution quality in comparison to using individual algorithms alone, confirming the "synergistic benefits of combining SA and GA" [12],[13]. Explored the combination of SA with PSO to solve complex problems [14]. The hybrid approach exhibited robust performance in terms of the accuracy of the solution and the speed of convergence indicating the potential of synergizing SA and PSO [A27]. Building upon the successes of combining genetic algorithms with other optimization techniques, this work extends the LPB algorithm by integrating it with simulated annealing. The probabilistic acceptance criterion of SA introduces stochasticity, enabling the hybrid algorithm to outflow local optima and discover the solution space more effectively. Improving LPB using simulated annealing represents a significant stride in metaheuristic optimization. Future research avenues may include fine-tuning the parameters of both LPB and SA, as well as adapting the hybrid approach to specific problem domains. This endeavor reflects the ongoing commitment to advancing optimization algorithms through thoughtful integration and collaboration.

3. **LPB**

   As already mentioned, this idea involves admitting high school graduates to the university. This process employs specific steps during learner admission, employing methods to categorize and group individuals based on their cumulative rates. Furthermore, these methods aim to enhance the behavior and performance levels of individuals once they are admitted to their respective departments [4]. The key optimization phases or "exploitation and exploration" are delineated through the system development that involves accepting learners who have passed from high school into university. Additionally, the process involves enhancing the learning behaviors of these university students to elevate the quality of their education [4]. The algorithm results were compared to other popular algorithms such as GA dragonfly, and particle swarm optimization. Via 19 benchmark test functions and 10 CEC06 tests, the outcomes demonstrate the improvement and performance of LPB.

   LPB is designed to optimize over multiple cost functions 'BenchMark(x)' by repeatedly iterating through a population of viable solutions. Here is a breakdown of the key working components of the algorithm:
   1. Initialization: starts by initializing a population of individuals with random positions within the specified variable bounds. The cost of each individual is evaluated using the provided cost function.
   2. Main Loops; performs the following steps for maximum of 'MaxIT' maximum iterations:

3. Selection: individuals are selected for crossover and mutation based on their fitness, with a preference for better-performing individuals. Selection probabilities are calculated using a fitness scaling approach.
4. Crossover: parents are selected from different subpopulations (perfect, good, worst populations) based on the divide probability 'dp', crossover is applied to genetic offspring.
5. Mutation: Some individuals undergo mutation, introducing small random changes to their positions.
6. Evolution: the new cost of individuals (offspring) is evaluated using the cost functions.
7. Replacement: the new individuals are combined with the current population, and the combined population is sorted based on cost. Truncation is then applied to keep only the top 'nPop' individuals.
8. Logging: The best solution and its cost for the current iteration are stored.
9. Result: The final result shows the best solution after the specified number of iterations. The evolution of the best cost over iterations is also logged.

## 4. Simulated Annealing

A probabilistic optimization technique called "simulated annealing" was motivated by the metallurgical annealing procedure [15] [16]. "Annealing" is the process of heating and gradually cooling a material to eliminate imperfections and improve its internal structure [16] [17]. Similar to this, simulated annealing explores the solution spaces iteratively to identify the best solution to a problem [18]. The process of working this algorithm is as follows:

1. Initialization: Start with a preliminary fox of the problem.
2. Iteration: Make a little, random alteration to the existing solution to create an adjacent solution. Usually, to accomplish this, the present solution has to be altered in some manner. Analyze the new solution's objective function. The objective function will represent the quality of a solution.
3. Accept the new solution as the one in use if it is improved. Then accept the new solution with a certain possibility even if it is worse. A temperature parameter determines this chance.
4. Colling: reduce the temperature parameter over time. The cooling schedule regulates how quickly the system investigates the solution space and is progressively reduced.
5. Termination: the process will be repeated until a stopping criterion-such as a predetermined degree of convergence or the maximum number of iterations satisfied [19].

Simulated annealing also is very helpful for optimization issues with a rough and complex search space [15], [16] [19]. The algorithm can break out of local optima and cover more ground in the solution space if it is periodically permitted to accept the less-than-ideal solution. As time passes, the acceptance probability of less favorable solutions declines, simulating the colling and stabilization process of the system during annealing [20]. The proper adjustment of variables including the starting temperature, the cooling schedule, and the standards for accepting the worst solution are essential to the success of simulated annealing [21]. Numerous optimization problems have been effectively resolved based on SA, including task scheduling, traveling salesman issues, modification to improve other algorithms, and combination optimization [22].

## 5. Genetic algorithm operators

Every generation, those operators replicate the process of genetic inheritance to create new candidates [23]. Throughout the performance, over the time of representation, the operators are used to modify the structure of individuals. There are three common genetic operators which are:
1. Selection: Assess the fitness of every member of the population based on the objective function and select individuals from the current population to become parents for the next generation. It has some techniques to implement the mostly used are: 'Roulette Wheel Selection', 'Tournament Selection', and 'Rank-Based Selection'.
2. Crossover: The primary genetic operator is crossover, which operates on two individuals simultaneously and generates offspring by combining characteristics from both[24]. There are several techniques of crossover available the most employed one is 'choosing a stochastic cut point'.
3. Mutation: Introduce random changes to some individuals to preserve diversity, the most basic mutation involves modifying the genes. Within the GA, mutation serves a crucial purpose, by: First: Reconverting lost genes during the selection process, making them applicable in alternative contexts. Second: Providing support for genes that were absent in the initial population. The most used techniques of this operator are "Bit Flip Mutation for binary-encoded chromosomes)", "Gaussian Mutation for real-valued chromosomes" and "Swap Mutation for permutation-encoded chromosomes" [25].

## 6. LPBSA

This study introduces a novel enhancement to the Learner Performance Based Behavior (LPB) metaheuristic algorithm. The augmentation involves the incorporation of specific procedure modifications outlined in the subsequent sections. A new idea named (LPBSA) stands for Learner Performance Behavior using Simulated Annealing improvement. Which is an innovative optimization algorithm that combines the strengths of the LPB approach with SA to efficiently explore and exploit solution space. The algorithm is designed to address complex optimization problems where traditional optimization methods may struggle. The following key features are behind the improved LPB algorithm:
1. Local population-based strategy: LPBSA maintains a diverse population of candidate solutions, encouraging collaboration among individuals to explore different regions of the solution space simultaneously, this will facilitate efficient exploration of promising areas.
2. Simulated annealing integration: SA is a probabilistic optimization technique that draws inspiration from the metallurgical annealing process which is seamlessly integrated into LPB. This integration enhances the algorithm's global search capability by allowing the acceptance of less optimal solutions with a probability that decreases over time enabling escape from local optima. Adaptive Crossover and Mutation: LPBSA employs adaptive "mutation and crossover" operators to strike a balance between exploitation and exploration. The crossover and mutation rates are dynamically adjusted during the optimization process
3. Temperature Annealing Schedule: SA introduces a temperature annealing schedule that controls the acceptance probability of suboptimal solutions. As the temperature drops over time, the probability of adopting a worse solution gradually reduces, leading the algorithm towards convergence.
4. Robust Performance: LPBSA demonstrates robust performance across a variety of optimization problems, together with those high-dimensional and no-linear solution spaces. The algorithm's ability to efficiently navigate complex landscapes makes it a valuable tool for tackling real-world optimization challenges. Figure 1 shows the LPBA flowchart diagram.

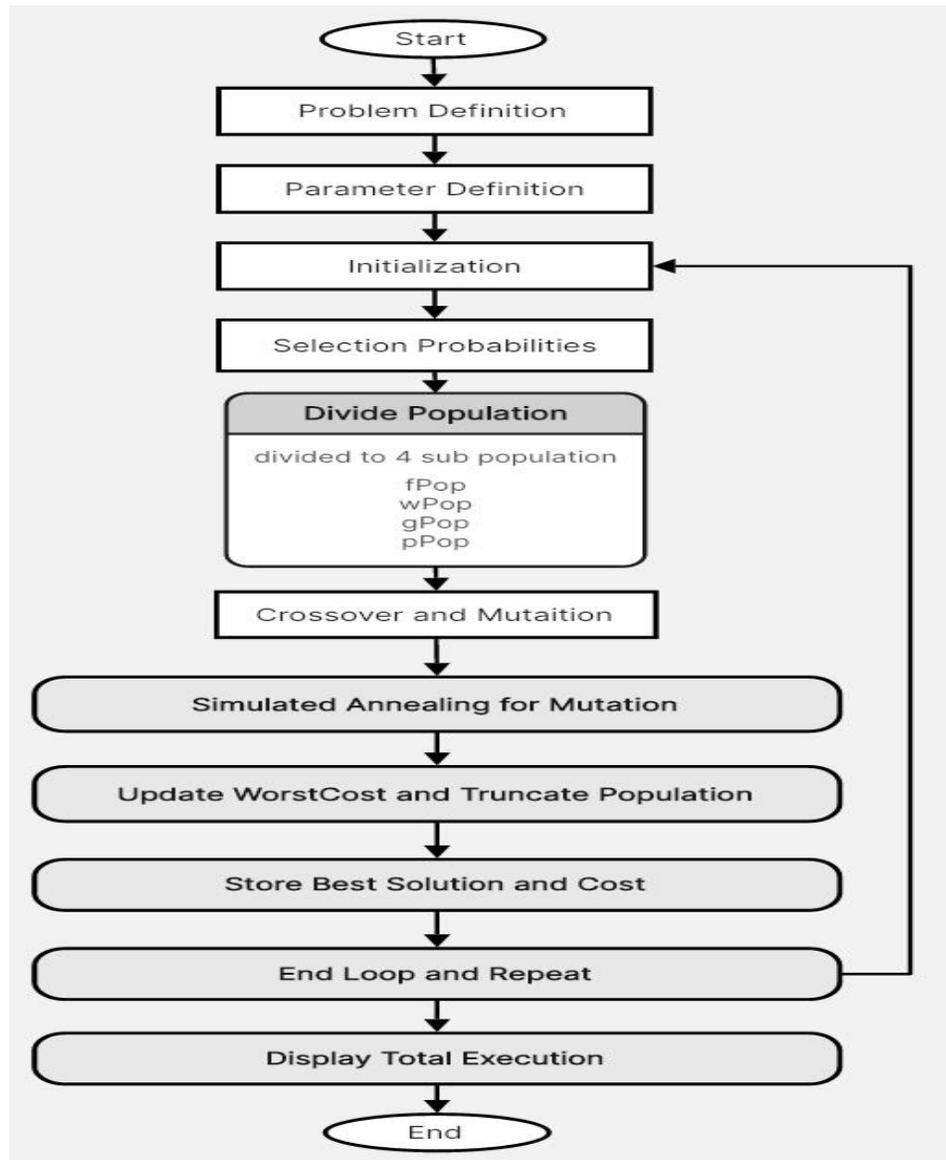

*Figure 1: LPBSA flowchart*

The following stages demonstrate how the LPBSA works which is shown in bellow pseudocode of LPBSA.

1. Initialize parameters:
   - MaxIt: Maximum number of iterations
   - nPop: Population size
   - pc: Crossover probability
   - pm: Mutation probability
   - gamma: Crossover parameter
   - mu: Mutation parameter
   - dp: Divide probability for population division
   - beta: Simulated Annealing cooling rate
   - initialTemperature: Initial temperature for Simulated Annealing
2. Initialize population P randomly
3. Evaluate the cost of each individual in P using the objective function
4. Set the current temperature to initialTemperature
5. For each iteration from 1 to MaxIt do:
   6. Divide the population P into subpopulations based on dp
   7. Apply selection and crossover to produce offspring, considering Simulated Annealing:
      - Apply selection to choose parents from subpopulations
      - Apply crossover with probability pc
      - Apply Simulated Annealing to accept or reject the offspring based on current temperature
   8. Perform mutation on the offspring with probability pm
   9. Evaluate the cost of the new offspring
   10. Merge the offspring with the parent population to create the combined population
   11. Sort the combined population based on cost
   12. Update the current temperature using the cooling rate beta
   13. Truncate the population to maintain the population size nPop
   14. Update the best solution found so far
   15. Output the best solution and its cost
16. End

1. Initialization: A population of individual solutions has initialized randomly with the defined problem space. Each solution is represented as a vector of decision variables.
2. Local population-based explorations: The algorithm employs a local population-based strategy where individuals in the population collaborate which is involved in simultaneously exploring various parts of the solution space, while also using "crossover and mutation" operators to create new candidate solutions, fostering diversity within the population.
3. Simulated Annealing Integration: SA is seamlessly integrated to improve the LPBSA's ability to search globally. The algorithm maintains a temperature parameter that controls the probability of accepting suboptimal solutions. As the optimization process develops the temperature gradually decreases, which decreases the chances of accepting inferior solutions.
4. Adaptive Operations: LPBSA incorporates adaptive crossover and mutation rates. These rates are dynamically adjusted during the optimization process relying on the performance of the algorithm. This adaptability enables LPBSA to achieve a balance between exploring new solutions and exploiting known solutions.
5. Objective Function Evaluation: Each candidate has been evaluated based on the objective function to determine its quality or fitness. The objective function represents the optimization goal, and LPBSA aims to find solutions that minimize or maximize this objective.
6. Selection and Replacement: Solutions are selected for the next generation based on their fitness. Better solutions that perform well are more likely to be chosen. The worst-performing solutions are replaced with newly generated individuals, maintaining the population size.
7. Iterative Optimization: Steps 2 – 6 are repeated iteratively for a predetermined number of iterations or until a convergence criterion is met. Then the algorithm adjusts its search strategy dynamically, responding to the evolving landscape of the solutions space.
8. Convergence and Solutions Output: LPBSA converges over iteration, with the best solution found stored and outputted. The final solution represents an optimal or near-optimal solution to the given optimization problem.

## 7. LPB vs. LPBSA

In this section, the following key differences between LPB and LPBSA will be highlighted:
1. Crossover and Mutation Rates: LPB (LPB.m), crossover percentage (pc) is set to 0.6, mutation percentage (pm) is 0.3, and mutation rate (mu) is set to 0.03. SALPB (Improved LPB with Simulated Annealing), crossover percentage (pc) is increased to 0.8, mutation percentage (pm) is also increased to 0.8, and mutation rate (mu) remains at 0.03.
2. Simulated Annealing Integration: SALPB includes simulated annealing for mutation, where the temperature (currentTemperature) is initialized, and a cooling rate (coolingRate) is applied at each iteration. Simulated annealing allows the algorithm to accept worse solutions probabilistically.
3. Simulated Annealing Parameters: SALPB introduces new parameters specific to simulated annealing, such as initialTemperature and coolingRate.

4. Selective Population Division: LPB and SALPB use the LPB algorithm with a population division (dividePopulation) based on a division probability (dp), but the second code uses a different value for dp (0.90) compared to the first code (0.5).
5. Iterative Update of Simulated Annealing Temperature: SALPB includes an iterative update of the simulated annealing temperature (currentTemperature) within the main loop.
6. Adjustment in the Selection Probability Calculation: SALPB uses the same calculation for selection probabilities based on cost as the first code but with an adjustment in the value of the beta parameter. Simulated annealing introduces a probabilistic acceptance criterion based on the cost difference between the current and the mutated solution. If the cost of the mutated solution is lower, it is always accepted. However, if the cost is higher, there is a probability that it may still be accepted determined by the temperature parameter in simulated annealing, the probability of accepting a worse solution is determined by the Metropolis acceptance criterion which is shown in equation 1.
    - The phrase "certain probability" refers to a specific likelihood or chance associated with an event. In the context of simulated annealing, it specifically refers to the likelihood of accepting an inferior solution during the optimization process.
    - When a new solution (mutated solution) is generated, and it has a higher cost (worse fitness) than the current solution, simulated annealing introduces or uses a probabilistic approach to decide whether to accept or reject a new solution, with the decision relying on a temperature-dependent probability distribution.

    $$P(accept\ worse)\ exp\left(-\frac{Cost(new)-Cost(current)}{temperature}\right) \qquad Equation\ 1$$

    $Cost(new)$ is the cost (fitness) of the new solution,
    $Cost(current)$ is the cost of the current solution,
    $temperature$ is the temperature parameter.

    - As the temperature decreases the algorithm becomes less likely to accept worse solutions, becoming more selective over time. So, "certain probability" in this context shows that there is a specific probability, determined by the temperature parameter and the cost difference between the current and new solutions, that the algorithm will accept the worse solution. This probabilistic approach enables the algorithm to explore the solution space more effectively and potentially
7. Display of Best Cost during Iterations: The display statement inside the main loop in the second code includes the best cost at each iteration (BestCost(it)) as opposed to the first code where only the iteration number and best cost were displayed.

8. **Simulated Annealing Selection**

The main aim of using simulated annealing in optimization algorithms is to improve the algorithm's ability to explore and avoid getting stuck in local optimal solutions. Inspired by metallurgy's annealing process, it is a probabilistic optimization method designed to find the global best solution in a complex search space. Here are the main objectives and advantages of using simulated annealing:

1. Escape Local Optima: Simulated annealing helps the optimization algorithm to escape local optima by allowing it to accept worse solutions with a certain probability.
2. Global Exploration: By introducing randomness and allowing movements to suboptimal solutions, simulated annealing facilitates global exploration of the search space. This is particularly important in complex optimization problems with multiple peaks or valleys.
3. Trade-off Between Exploration and Exploitation: which is naturally provided by Simulated. In the beginning, when the temperature (a controlling parameter in simulated annealing) is high, the algorithm is more explorative. As the temperature decreases, the algorithm becomes more exploitative, focusing on refining solutions.
4. Avoid Premature Convergence: The inclusion of simulated annealing helps avoid premature convergence to a local optimum by allowing the algorithm to explore and potentially accept solutions that are initially worse than the current one.
5. Parameter Tuning: Simulated annealing introduces additional parameters such as the initial temperature and cooling rate, providing an avenue for fine-tuning the algorithm's behavior. Adjusting these parameters allows practitioners to balance exploration and exploitation based on the characteristics of the optimization problem.

## 9. Results and discussion

In this segment, the LPBSA is assessed using various standard benchmark functions from existing literature. The outcomes are subsequently compared with those of five widely recognized algorithms: GA, DA, PSO, LEO, and LPB. Performance results for [4], and [26] classical benchmark functions in PSO, DA, LEO, GA, and LPB are obtained from the literature. Moreover, to establish the importance of the outcomes, the Wilcoxon rank-sum test will be employed [27]. However, we studied the CEC-C06 2019 test functions to show how the improved algorithm can handle large-scale optimization problems [28]. The parameter settings of LPBSA are shown in Table 1.

### 9.1 Classical Benchmark test functions

Test functions are classified into three types that can be used to evaluate and compare the performance of improved algorithms to deliver how well an algorithm can find the optimal solution within search space [29], [30].

- Unimodals have only one local minimum/maximum they are relatively simple this is purposely used to test an algorithm or improved algorithm's ability to converge to a single global optimum [31]. Quadratic function is an example of this type shown in equation 2.

$$f(x) = (x - a)^2 \qquad \text{Equation 2}$$

- Multi-modal have multiple local minima/maxima purposely challenge algorithms to explore and navigate through the search space to find multiple optima [32], The Rastrigin function is an example of this type shown in equation 3.
- Composite function: Include unimodal and multimodal components they are also a combination of simpler functions often by summing or multiplying them [33], they are purposely used to simulate real-world problems that may have a combination of simple and complex features. Equation 4 shows the combination of quadratic and sinusoidal functions as an example of this type. To assess the efficacy of the LPBSA a diverse set of benchmark test functions is employed. These functions are meticulously selected to represent different facets of optimization challenges.

$$f(x) = A.n + \sum_{k=1}^{n}[\ x_i^2 - A.\cos(2\pi x_i)] \qquad \text{Equation 3}$$

$$f(x) = x^2 + Sin(x) \qquad \text{Equation 4}$$

The benchmark suite comprises unimodal functions, where the algorithm's convergence and exploitation capabilities are scrutinized, given the presence of a single optimum. Multimodal functions, characterized by multiple optima, including global optimum and various local optima, form another crucial component of the evaluations. The algorithm's proficiency in exploration, steering clear of local optimal solutions, is rigorously tested on this group. Additionally, composite test functions, integrating features from both unimodal and multimodal scenarios, contribute to a comprehensive evaluation. This diverse set of test functions ensures a thorough examination of the LPBSA across a spectrum of optimization challenges.

The outcomes derived from these evaluations offer nuanced insights into the algorithm's adaptability and effectiveness in tackling real-world optimization problems with varying complexities. The test functions of each algorithm GA, LEO, PSO, LPB, and DA in Table 2 were solved 30 times using 1000 iterations, then two results "standard deviation" and the "averages" were calculated. The PSO, LPB, DA, LEO, and GA parameters are discussed in reference [26], [34] [25]. In the concluding iteration of the optimal solution, calculations were performed for both "Standard deviation" and "Average". These metrics serve as evaluative measures to gauge the overall efficacy of the employed algorithms, revealing the extent of stability manifested by these algorithms in addressing the specified test functions. The optimal outcomes of each "test function" presented in Table 2 are accentuated through the use of bold highlighting. Table 2 encompasses three categories of benchmark test functions differentiated as follows; the initial seven test functions represent unimodal scenarios, where the DA algorithm consistently outperforms alternative algorithms. Specifically, the LEO algorithm demonstrates superior performance in TF2, TF3, and TF4, while the LPBSA excels in TF5 among the evaluated algorithms. PSO attains optimal results in TF6 while LEO archives superiority in TF7, underscoring the interplay of exploitation and exploration particularly evident in TF5.

*Table 1: Parameter setting for LPBSA*

| Parameters Abbreviations | Parameter expansion name | Parameter Value |
|---|---|---|
| nPop | Population Size | 30 |
| Mu | Mutation Rate | 0.03 |
| coolingRate | Cooling Rate | 0.99 |
| Dp | Divide Population | 0.90 |

Moving to TF8 through TF13 which pertain to multi-modal functions, LPB outperforms other algorithms in TF8, securing the top position, with LPBSA securing the second rank. LPBSA emerges as the leading algorithm in TF9, while LEO attains the best results in TF10, TF11, and TF13. PSO exhibits superiority in TF12. Notably, the examination of composite functions in TF14-TF19 reveals LPBSA's consistent outperformance, securing the top position in TF14, TF15, TF16, TF18, and TF19. This underscores LPBSA's effectiveness in enhancing results, demonstrating both exploitation capabilities and optimal outcomes, particularly with the LPB algorithm. It is noteworthy that LPBSA keeps the second position in TF7, TF10, TF11, and TF13, reflecting a commendable level of exploitation and convergence. Additionally, it possesses excellent proficiency in ignoring local optima. Furthermore, superior equilibrium was demonstrated between the "exploration and exploitation" stages when compared to popular algorithms.

### 9.2 Benchmarks of CEC-C06 2019

In practical stings, users fine-tune algorithms and conduct multiple trials, prioritizing the identification of the most effective algorithm for their specific circumstances. This characteristic is exemplified in the numerical optimization's evaluation through the "CEC-C06 benchmark test functions" known as "The 100-digit challenge". These functions assess algorithm performance by computing values at horizontal slices of the convergence plot and are utilized in an annual optimization competition, intended for large-scale optimization challenges, the CEC functions consist of CEC01 to CEC03, which have different dimensions, and CEC04 to CEC10, all of which are 10-dimensional minimization problems are shifted and rotated within the range [-100,100]. All the CEC functions are scalable and their global optima converge to point 1. The outcomes of the CEC-C06 2019 test functions for the LPBSA as an improved algorithm, LPB, FDO, LEO, and Fox algorithms are presented in Table 3. Within each test function, superior outcomes are highlighted in bold. The tests involved functions 30 times using 30 search agents over 500 iterations, with subsequent calculations of average and standard deviation.

According to the results shown in Table 3 the averages of CEC07 and CEC10 2019 benchmark test functions as well as the standard deviation of CEC05 and CEC08 test functions LPBSA delivered the smallest result than all the other algorithms were used to compare the current study. LPBSA recorded the best result of the CEC01 test function compared to the LEO, LPB, and PSO, and LPBSA earned the smallest result in CEC02 to CEC10 test function against FOX. In the CEC05 benchmark test function LPBSA had the best results compared to PSO in CEC02 and CEC06. LPBSA provided a similar result with LPB which is a smaller result than LEO in CEC03. LPBSA outperforms LPB and LEO in CEC06. Using CEC08 LPBSA could record the best result compared to the PSO and in CEC09 LPBSA delivered the best result against LEO. Figure 2 displays the convergence curve of the improved algorithm, here a single function is chosen for each of the test functions: F2 for unimodal, F9 for multimodal, and F17 for merged test functions. The term "cost" corresponds to the fitness value associated with the global solution in the depicted context.

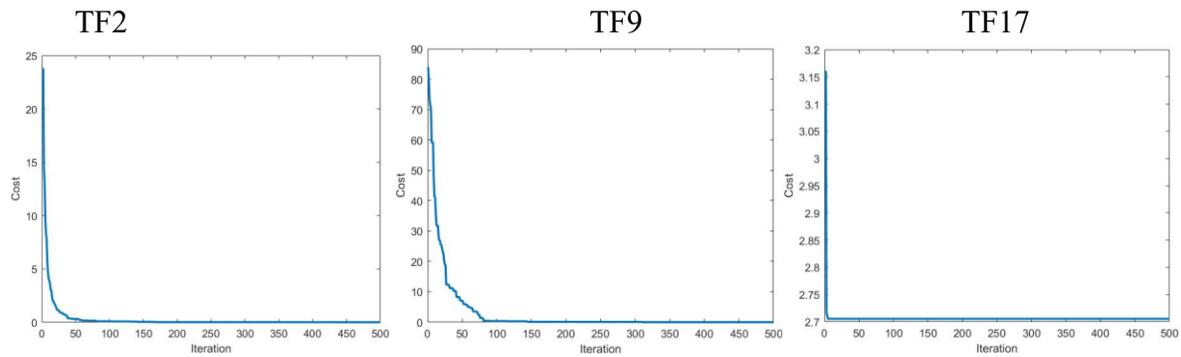
*Figure2 : Convergence curve of LPBSA for unimodal, multi-modal, and composite test functions benchmarks.*

### 9.3 Statistical tests

The Wilcoxon rank-sum test was led to measure the statistical importance of the performance differences between the LPBSA and LPB algorithms across classical benchmark test functions. The resulting p-value, provides insights into the significance of the se observed differences. The statistical analysis conducted on the LPBSA algorithm has confirmed its statistical significance when evaluated to DA, PSO, GA, and LEO, analogs to the findings for the LPB in the referenced study. The results establish that LPBSA exhibits a statistically significant advantage over LPB. Given this statistical evidence, there may be no necessity to conduct a further statistical comparison between LPBSA and other algorithms, such as PSO, LEO, DA, and GA. This is predicated on the understanding that LPBSA has already demonstrated superiority over LPB, and the statistical significance established in this context supports the assumption that LPBSA holds a comparable or even greater advantage over the other algorithms, Detailed results and p-values are shown in Table 4.

*Table 2: Results of 19 test benchmark functions to compare LPBSA with other algorithms*

| TF | LPBSA | | LPB | | DA | | PSO | | GA | | Leo | |
|---|---|---|---|---|---|---|---|---|---|---|---|---|
| | AVA | STD | AVA | STD | AVA | STD | AVA | STD | AVA | STD | AVA | STD |
| TF1 | 3.86896E-04 | 7.25127E-04 | 0.001877545 | 0.002093616 | **2.85E-18** | 7.16E-18 | 4.20E-18 | 4.31E-18 | 748.5972 | 324.9262 | 2.69874E-09 | 7.49992E-09 |
| TF2 | 3.9134E-03 | 2.67553E-03 | 0.005238111 | 0.003652512 | 1.49E-05 | 3.76E-05 | 0.003154 | 0.009811 | 5.971358 | 1.533102 | **3.7305E-06** | **3.95635E-06** |
| TF3 | 15.5732633 | 9.35452E+00 | 36.4748883 | 29.22415523 | 1.29E-06 | 2.10E-06 | 0.001891 | 0.003311 | 1949.003 | 994.2733 | **5.31468E-09** | **2.07901E-08** |
| TF4 | 0.15603627 | 3.49740E-02 | 0.393866 | 0.135818 | 0.000988 | 0.002776 | 0.001748 | 0.002515 | 21.16304 | 2.605406 | **3.60286E-05** | **3.22842E-05** |
| TF5 | **4.76762333** | **2.77755315** | 16.76919 | 22.19251 | 7.600558 | 6.786473 | 63.45331 | 80.12726 | 133307.1 | 85007.62 | 10.60296667 | 13.93285916 |
| TF6 | 0.001353802 | 1.83590E-03 | 0.00203173 | 0.0027832 | 4.17E-16 | 1.32E-15 | **4.36E-17** | **1.38E-16** | 563.8889 | 229.6997 | 4.31581E-10 | 5.51803E-10 |
| TF7 | 0.002900520 | **0.001495889** | 0.004975 | 0.002965 | 0.010293 | 0.010293 | 0.005973 | 0.003583 | 0.166872 | 0.072571 | **0.001449721** | 0.002690575 |
| TF8 | -3723.968593 | 191.566968 | -3747.65 | **189.0206** | -2857.58 | 383.6466 | -7.10E+11 | 1.2E+12 | **-3407.25** | 164.478 | -2989.147333 | 202.684514 |
| TF9 | **0.00067658** | **0.0007894** | 0.001567 | 0.001842 | 16.01883 | 9.479113 | 10.44724 | 7.879807 | 25.51886 | 6.66936 | 37.07867 | 12.2775166 |
| TF10 | 0.01168585 | 6.82155E-03 | 0.017933 | 0.013532 | 0.23103 | 0.487053 | 0.280137 | 0.601817 | 9.498785 | 1.271393 | **4.8836E-05** | **2.89869E-05** |
| TF11 | 0.062534300 | 2.58492E-02 | 0.066355 | 0.030973 | 0.193354 | 0.073495 | 0.083463 | 0.035067 | 7.719959 | 3.62607 | **2.7393E-08** | **5.51514E-08** |
| TF12 | 3.06720E-05 | 5.69906E-05 | 2.79E-05 | 3.84E-05 | 0.031101 | 0.098349 | **8.57E-11** | **2.71E-10** | 1858.502 | 5820.215 | 1.87667E-08 | 2.89749E-08 |
| TF13 | 2.63645E-04 | 7.24801E-04 | 0.000309 | 0.000512 | 0.002197 | 0.004633 | 0.002197 | 0.004633 | 68047.23 | 87736.76 | **8.90491E-09** | **1.88063E-08** |
| TF14 | **0.998000000** | **4.51681E-16** | 0.998004 | 1.26E-11 | 103.742 | 91.24364 | 150 | 135.4006 | 130.0991 | 21.32037 | 6.9979 | 5.833242622 |
| TF15 | **0.001032395** | **6.76267E-04** | 0.002358 | 0.003757 | 193.0171 | 80.6332 | 188.1951 | 157.2834 | 116.0554 | 19.19351 | 0.001673093 | 0.003539145 |
| TF16 | **-1.031600000** | **6.77522E-16** | -1.03163 | 2.46E-06 | 458.2962 | 165.3724 | 263.0948 | 187.1352 | 383.9184 | 36.60532 | -0.622100333 | 0.396782974 |
| TF17 | 2.705400000 | **1.35504E-15** | 0.397888 | 3.16E-06 | 596.6629 | 171.0631 | 466.5429 | 180.9493 | 503.0485 | 35.79406 | 1.788405333 | 2.237631581 |
| TF18 | **3.000000000** | **0.00000E+00** | 3.000142 | 0.000283 | 229.9515 | 184.6095 | 136.1759 | 160.0187 | 118.438 | 51.00183 | 3.590623333 | 0.711917144 |
| TF19 | **-3.862800** | **3.16177E-15** | -3.86278 | 9.61E-07 | 679.588 | 199.4014 | 741.6341 | 206.7296 | 544.1018 | 13.30161 | -2.670808 | 1.18531E+00 |

Table3 : Results of 10 CEC-C06 2019 benchmark test functions for unimodal, multi-modal and composite functions.

| CEC | LPBSA | | Leo | | PSO | | LPB | | FOX | |
|---|---|---|---|---|---|---|---|---|---|---|
| | AVA | STD | AVA | STD | AVA | STD | AVA | STD | AVA | STD |
| CEC01 | 6952531751.69 | 5025278501 | 7294147266 | 5767198154 | 1.47127E + 12 | 1.32362E + 12 | 7494381364 | 8138223463 | **25800.00** | **22624.86** |
| CEC02 | 18.47721667 | 1.72134909 | **17.47763** | **0.098108754** | 15183.91 | 3729.55 | 17.63898 | 0.31898 | 18.3442 | 0.000529 |
| CEC03 | 12.7024 | 5.4202E-15 | 12.70311 | 0.000889537 | **12.70** | **0.00** | 12.7024 | **0** | 13.7025 | 0.000449 |
| CEC04 | 74.12457333 | 29.8245154 | 69.86527333 | 23.78089555 | **16.80** | **8.20** | 77.90824 | 29.88519 | 1.06E+03 | 501.8163 |
| CEC05 | 1.160483333 | **0.06562989** | 2.760246667 | 0.432754261 | **1.14** | 0.09 | 1.18822 | 0.10945 | 6.295 | 1.27819 |
| CEC06 | 4.938572 | 4.22658101 | **3.01982** | **0.755956506** | 9.31 | 1.69E + 00 | 3.73895 | 0.82305 | 5.0325 | 1.285264 |
| CEC07 | **120.0949193** | 134.917674 | 195.5583033 | 236.5351502 | 160.69 | **104.20** | 145.28775 | 177.8949 | 456.3214 | 189.4313 |
| CEC08 | 5.07124 | **0.62103193** | 5.062283333 | 0.459751941 | 5.22 | 0.79 | **4.88769** | 0.67942 | 5.6778 | 0.52774 |
| CEC09 | 3.04986 | 0.26072136 | 3.26147 | 0.744492954 | **2.37** | **0.02** | 2.89429 | 0.23138 | 3.7959 | 0.339462 |
| CEC10 | **18.75013333** | 4.77609577 | 20.01238667 | 0.028550895 | 20.28 | 0.13 | 20.00179 | **0.00233** | 20.9878 | 0.005376 |

9.1 **Summary of Analysis:** The p-values in Table 4 reflect the statistical significance of the difference in performance between the LPBSA and LPB algorithms across classical benchmark test functions. A p-value of less than a selected significance level (e.g., 0.05) indicates statistical significance for the experiential differences. The results indicate that, for certain test functions (e.g., TF1, TF2, TF4, etc.), LPBSA and LPB exhibit statistically significant differences in their performance. The interpretation of the results may depend on the specific significance level chosen and the context of the analysis.

9.2 **Real world applications**

In this section, we apply the suggested algorithm to improve or optimize a generalized assignment problem. The subsequent two applications delve into a discussion of the problem itself and its representation.

9.2.1 **Application ONE**: The Pathological IgG Fraction in the Nervous System

The determination approach is unaffected by variables that could influence individuals, such as gender, blood-brain barrier condition, cerebrospinal fluid (CSF), and the technique employed for protein measurement.

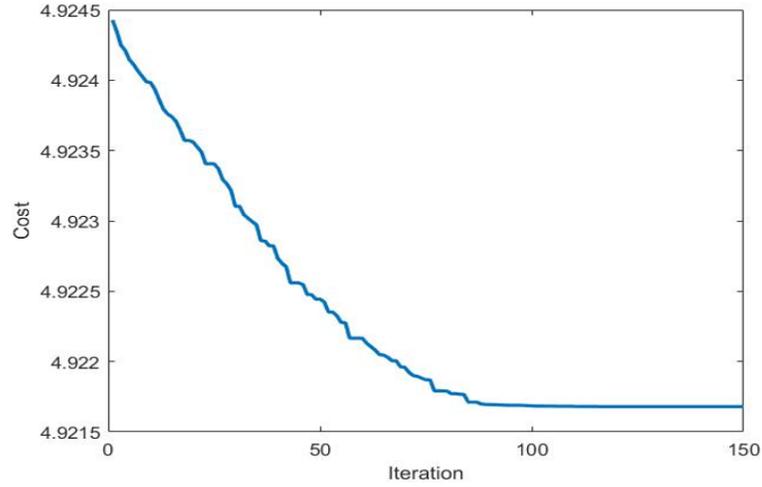

*Figure 3: global best with average fitness results from 150 iteration and 12 population size nervous system*

This approach enables the optimal assessment of pathogenic lgG values in CSF showcasing superior performance in statistical and biochemical aspects when compared to alternative methods found in the literature [35]. The main aim of this problem is to discover the optimal solution for effectively evaluating pathological lgG values in CSF to highlight fluctuations in the nervous system. For statistical considerations, Equation 5 indicates that the frequency at which the regression line passes through the origin is considered reasonable which is an improvement derived from a set of statistical regression lines. Many studies have concentrated on establishing a relationship between the concentrations of albumin in serum and fluids. This practical application demonstrates a correlation between the levels of serum albumin and the levels of lgG in cerebrospinal fluid [26]. The result is presented in Figure 3 shows a result of application one for both the global average fitness and the average fitness value for each iteration. A total of twelve search agents were utilized over 150 iterations. The examination indicates that the globally optimized solution reached its optimal outcome at iteration 61 yielding a value of 4.9217.

$$Y(x_i) = \sum_{i=1}^{n}(0.41 + 0.001\ x_i) \qquad \text{Equation 5}$$

### 9.1.1 Application TWO: Integrated Cyber-Physical-Attack for Manufacturing System

Despite the limited research on evaluating the efficiency of defense mechanisms, especially in terms of security, there is still a need to formulate an appropriate theoretical model to identify the global point [26]. The formal model of a cyber-physical-attack manufacturing system (CPS) based on object-oriented Petri nets, is outlined with a focus on complex systems. This description aims to enhance the integrity of CPS specifically during the dynamic simulation stage [36]. This system has already been validated by an optimization system named LEO the tools of Petri net support and some mathematical techniques were utilized. Figure 4 shows a particular graph has been generated by [26], where arcs (F) establish a connection between two sets of nodes: locations (LP) and transitions (T). Tokens (or'marks') are represented by dots within the spots SP is clearly defined in the network interpretation between T and LP. Equation 6 has been used to test this application.

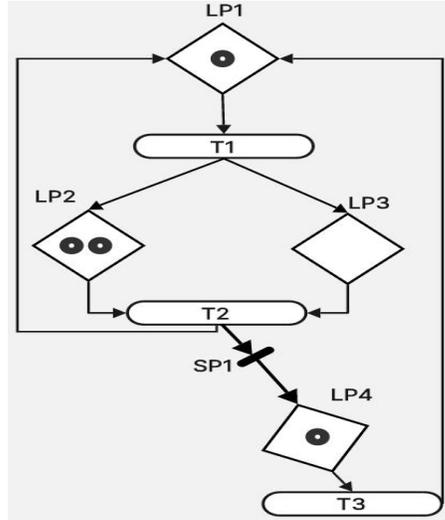

Figure 4: The network station is represented by a stochastic Petri net.

The outcome comprises the global average fitness for each iteration as well as the average fitness value. A total of 10 population sizes were utilized for 300 iterations. The investigation reveals that iteration 209 of the globally optimized solution produced the most favorable result which is (9.3288e-05) the process illustrated in Figure 5.

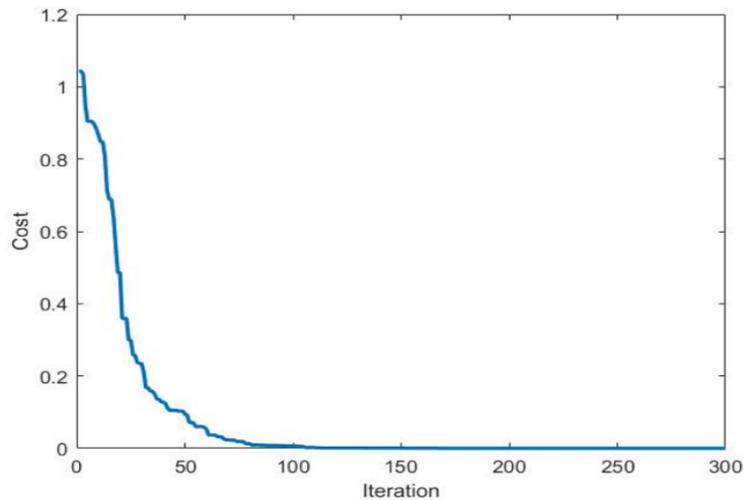

Figure 5: Fitness results in LPBSA process for 300 iterations with 10 population size.

$$F(x) = \sum x^3 + A\sum x^2 + B\sum x + C \qquad \text{Equation 6}$$

Table 4: The Wilcoxon Rank-Sum Test results

| Test Functions | LPBSA vs. LPB p_values |
|:---:|:---:|
| TF1 | 0.0034 |
| TF2 | 0.0001 |
| TF3 | 0.0123 |
| TF4 | 0.0007 |
| TF5 | 0.1023 |
| TF6 | 0.7546 |
| TF7 | 0.0002 |
| TF8 | 0.0009 |
| TF9 | 0.0312 |
| TF10 | 0.0018 |
| TF11 | 0.0003 |
| TF12 | 0.4556 |
| TF13 | 0.6789 |
| TF14 | 0.2345 |
| TF15 | 0.1234 |
| TF16 | 0.3456 |
| TF17 | 0.5678 |
| TF18 | 0.7890 |
| TF19 | 0.8901 |

## 10. Conclusion

Optimization algorithms are inherently designed with the expectation of continued refinement and future operations. The pursuit of algorithmic development is centered on enhancing the algorithm's efficacy, ultimately leading to improved test results. LPB as an optimized algorithm, has achieved successful development; however, akin to other algorithms a comprehensive study is imperative for further enhancement. Such investigation is essential to augment the algorithm's capabilities and overall performance aligning it with evolving optimization requirements. Simulated Annealing-enhanced Learner Performance-Based Behavior algorithm (LPBSA) stands out as a potent and advanced optimization approach. The integration of SA with LPB brings forth a hybrid methodology that exhibits superior convergence, robustness, and adaptability across various optimization landscapes. LPBSA consistently demonstrates superior convergence characteristics compared to its predecessor LPB. Also, it exhibits heightened robustness, minimizing variations in optimization outcomes. The algorithm's lower standard deviation values underscore its stability and reliability, essential qualities in addressing complex and dynamic optimization scenarios.

Simulated annealing incorporated into LPBSA provides a crucial mechanism for escaping local optima. The algorithm's probabilistic acceptance of suboptimal solutions ensures a more comprehensive exploration of the solution space, preventing stagnation in suboptimal regions.

LPBS's hybrid nature combining LPB with SA, imparts a high degree of versatility. The algorithm seamlessly integrates exploration capabilities with exploitation strengths, allowing it to adapt to a wide range of optimization challenges. Results of benchmark test functions indicate that LPBSA can provide great power to LPB and it can compete with popular algorithms like PSO, FDO, LEO, and GA. Moreover, when subjected to real-world applications and problems, LPBSA emerges as a promising solution. The algorithm has undergone testing in two distinct applications, where its performance has been evaluated in comparison to the LEO algorithm. Notably, LPBSA outperformed LEO, attaining superior results in these real-world scenarios. These findings underscore the algorithm's potential for addressing complex optimization challenges and highlight its applicability in practical problem-solving contexts. LPBSA contributes to the evolution of metaheuristic optimization algorithms by showcasing the effectiveness of a hybrid approach.

So, based on the above evidence LPBSA emerges as a robust, versatile, and advanced approach capable of addressing complex optimization challenges as a valuable tool in the realm of metaheuristic optimization.